\documentclass[a4paper,fleqn]{cas-sc}

\usepackage[authoryear,longnamesfirst]{natbib}
\newtheorem{definition}{Definition}

\def\tsc#1{\csdef{#1}{\textsc{\lowercase{#1}}\xspace}}
\tsc{WGM}
\tsc{QE}
\tsc{EP}
\tsc{PMS}
\tsc{BEC}
\tsc{DE}
\begin{document}
\let\WriteBookmarks\relax
\def\floatpagepagefraction{1}
\def\textpagefraction{.001}

% Short title
\shorttitle{Assessing Markov Property in Driving Behaviors: Insights from Statistical Tests}

% Short author
\shortauthors{Li et~al.}

% Main title of the paper
% \title [mode = title]{This is a specimen $a_b$ title}      
\title [mode = title]{Assessing Markov Property in Driving Behaviors: Insights from Statistical Tests}
\author[1]{Zheng Li}[
orcid=0000-0001-5732-5605
]
% % Footnote of the first author
% \fnmark[1]
% Email id of the first author
\ead{zli2674@wisc.edu}

% % URL of the first author
% \ead[url]{www.cvr.cc, cvr@sayahna.org}

% %  Credit authorship
% \credit{Conceptualization of this study, Methodology, Software}

% Address/affiliation
\affiliation[1]{organization={Department of Civil and Environmental Engineering, University of Wisconsin-Madison},
    % addressline={Radarweg 29}, 
    city={Madison},
    % citysep={}, % Uncomment if no comma needed between city and postcode
    postcode={WI 53706}, 
    % state={},
    country={United States}}

% Second author
\author[2]{Haoming Meng}
\ead{hmeng29@wisc.edu}

% Address/affiliation
\affiliation[2]{organization={Department of Computer Science, University of Wisconsin-Madison},
    % addressline={Radarweg 29}, 
    city={Madison},
    % citysep={}, % Uncomment if no comma needed between city and postcode
    postcode={WI 53706}, 
    % state={},
    country={United States}}

% Third author
\author[1]{Chengyuan Ma}
% Corresponding author indication
\cormark[1]
\ead{cma97@wisc.edu}

% Fourth author
\author[1]{Ke Ma}
\ead{kma62@wisc.edu}

% Fifth author
\author[1]{Xiaopeng Li}
% Corresponding author indication
\cormark[1]
\ead{xli2485@wisc.edu}

% % Address/affiliation
% \affiliation[2]{organization={Sayahna Foundation},
%     % addressline={}, 
%     city={Jagathy},
%     % citysep={}, % Uncomment if no comma needed between city and postcode
%     postcode={695014}, 
%     state={Trivandrum},
%     country={India}}

% % Fourth author
% \author%
% [1,3]
% {Rishi T.}
% \cormark[2]
% \fnmark[1,3]
% \ead{rishi@stmdocs.in}
% \ead[URL]{www.stmdocs.in}

% \affiliation[3]{organization={STM Document Engineering Pvt Ltd.},
%     addressline={Mepukada}, 
%     city={Malayinkil},
%     % citysep={}, % Uncomment if no comma needed between city and postcode
%     postcode={695571}, 
%     state={Trivandrum},
%     country={India}}

% Corresponding author text
\cortext[cor1]{Corresponding authors: Chengyuan Ma, Xiaopeng Li}
% \cortext[cor2]{Xiaopeng Li}

% % Footnote text
% \fntext[fn1]{This is the first author footnote. but is common to third
%   author as well.}
% \fntext[fn2]{Another author footnote, this is a very long footnote and
%   it should be a really long footnote. But this footnote is not yet
%   sufficiently long enough to make two lines of footnote text.}

% % For a title note without a number/mark
% \nonumnote{This note has no numbers. In this work we demonstrate $a_b$
%   the formation Y\_1 of a new type of polariton on the interface
%   between a cuprous oxide slab and a polystyrene micro-sphere placed
%   on the slab.
%   }

% Here goes the abstract
\begin{abstract}
The Markov property serves as a foundational assumption in most existing work on vehicle driving behavior, positing that future states depend solely on the current state, not the series of preceding states. This study validates the Markov properties of vehicle trajectories for both Autonomous Vehicles (AVs) and Human-driven Vehicles (HVs). A statistical method used to test whether time series data exhibits Markov properties is applied to examine whether the trajectory data possesses Markov characteristics. \(t\) test and \(F\) test are additionally introduced to characterize the differences in Markov properties between AVs and HVs. Based on two public trajectory datasets, we investigate the presence and order of the Markov property of different types of vehicles through rigorous statistical tests. Our findings reveal that AV trajectories generally exhibit stronger Markov properties compared to HV trajectories, with a higher percentage conforming to the Markov property and lower Markov orders. In contrast, HV trajectories display greater variability and heterogeneity in decision-making processes, reflecting the complex perception and information processing involved in human driving. These results have significant implications for the development of driving behavior models, AV controllers, and traffic simulation systems. Our study also demonstrates the feasibility of using statistical methods to test the presence of Markov properties in driving trajectory data.
\end{abstract}

% Use if graphical abstract is present
% \begin{graphicalabstract}
% \includegraphics{figs/grabs.pdf}
% \end{graphicalabstract}

% Research highlights
\begin{highlights}
\item This work pioneers the application of statistical test methods to validate Markov properties in vehicle trajectory data, establishing a rigorous framework for analyzing driving behavior characteristics.
\item Statistical analysis reveals that autonomous vehicle trajectories demonstrate stronger Markov properties with lower Markov orders, demonstrate a higher percentage of trajectories conforming to the Markov property, and exhibit higher consistency in Markov order distributions compared to human-driven vehicles, reflecting fundamental differences in their decision-making processes.
\item Our findings provide crucial insights for developing more accurate driving behavior models and autonomous vehicle controllers and simulators by quantifying the extent to which the Markov assumption holds in different driving scenarios.
\end{highlights}

% Keywords
% Each keyword is seperated by \sep

\maketitle

\section{Introduction}
\par Markov property is a fundamental assumption in modeling and analyzing road vehicle driving behaviors. The property posits that in a sequential decision-making process, the decision at any given time is solely dependent on the current state and is independent of the history of previous states (\cite{chen2010markov}). A sequential decision-making process that satisfies the Markov property is termed a Markov decision Process (MP). Given this definition, we can infer that driving behaviors conforming to the Markov property is characterized by the vehicle's action being determined exclusively by its current kinematic state variables, including its motion state and surrounding traffic conditions, without dependence on its historical actions or states. Such property assumes a memoryless driving process, simplifying the inputs and models for driving behavior analysis, modeling, and simulation.

% \textcolor{red}{The Markov property has been widely adopted in existing research, particularly in car-following scenarios. Most car-following models rely solely on current state variables(e.g., ego speed, relative speed, and distance to the preceding vehicle), to determine the next-step decision (e.g., longitudinal acceleration rate). Classical and widely used models, including the Newell model, Gipps model, and IDM model, though differing in their modeling approaches, all assume the Markov property. Furthermore, the calibration of their key parameters is based on fitting driving behavior data under this assumption. As the foundation of traffic analysis, car-following models are extensively applied in various domains, including vehicle trajectory prediction, traffic flow modeling, traffic simulation, and traffic safety analysis. These applications often adopt and extend the Markov property assumption to simplify and enhance the modeling of vehicle driving behaviors. For example, xxxx. 

%  With the development of Autonomous Vehicles (AVs), research on AV testing and modeling continues to widely adopt this assumption, serving as a foundation for evaluating AV behaviors and their interactions in traffic systems.  xxx}
 
% \rule{\textwidth}{01pt}

\par Numerous past studies in this area have adopted the Markov property. For example, in studies focusing on modeling and analyzing human driving behaviors, the relationship between vehicle decisions and states is investigated based on the Markov property. The most representative examples are car-following models, e.g., the Newell model (\cite{newell2002simplified}), Gipps model (\cite{ciuffo2012thirty}), and Intelligent Driver Model (IDM, \cite{5550943}). Though differing in their modeling approaches, all assume the Markov property. Besides, the calibration of key parameters in these car-following models is based on fitting trajectory data under this assumption.

% Chen et al.\cite{chen2010markov} proposed a Markov car-following model that integrates headway distribution and vehicle interaction models to explain empirical headway distributions. Meng et al.\cite{meng2024car} created a jerk-constrained acceleration Markov car-following model, incorporating unconscious following and active acceleration modes. Kamrani et al.\cite{kamrani2020applying} applied a MP framework to analyze driving behaviors related to acceleration, deceleration, and speed maintenance decisions. Chion et al.\cite{chion2019managing} analyzed aggressive driving behaviors through a Markov model by considering short- and medium-term alternative measures. 

\par In studies working on trajectory prediction, they always predict the future position and speed of vehicles based on current states, which are also adoptions of the Markov property. For examples \cite{rathore2019scalable} proposed a scalable clustering and Markov-based framework for short- and long-term trajectory prediction. \cite{shin2018vehicle} proposed a Markov-based vehicle speed prediction algorithm.

\par Additionally, in work on Autonomous Vehicles (AVs) safety test, the Markov property is adapted to generate the trajectories for background Human-driven Vehicles (HVs). \cite{feng2023dense} used MP in the dense deep reinforcement learning approach to train the background agents. The MPs were revised by removing non-safety critical states and reconnecting critical ones to densify the information in the training data. Similar properties and assumptions were presented in other work (\cite{yan2023learning,feng2021intelligent}).

% \rule{\textwidth}{01pt}
% \textcolor{red}{Although most driving behavior modeling studies adopt the Markov property assumption and achieve reasonable accuracy, the influence of historical states on driving behaviors cannot be overlooked in certain scenarios, particularly in studies focusing on corner cases related to driving safety analysis. In fact, driving behaviors are intrinsically multifaceted, with numerous factors influencing the decision-making process and subsequent actions.  xxx}

\par However, some studies have emphasized drawbacks of the model paradigm based on the Markov property. The driving behaviors are intrinsically multifaceted, with numerous elements impacting the decision-making process and the consequent behaviors. Several studies have verified that incorporating not only the current states but also a portion of the historical states yields more accurate predictions for the driving behaviors. \cite{wang2017capturing} established a deep learning-based car-following model and proved that combining the past 10 seconds of the states can yield the most accurate prediction for future actions. \cite{pei2016empirical} proposed a car-following model with a gamma-distributed memory effect, which showed promising model accuracy. \cite{zhang2023calibrating} improved the fitness for the Bayesian Dynamic Regression car-following model by incorporating historical information at different scales. Thus, it is suggested that the Markov property does not always provide the most accurate description of driving behaviors. It is crucial to verify the validity of this assumption for vehicle trajectories of interest in driving behavior-related research.

\par Few existing studies have validated the Markov property. As mentioned before, the vast majority of the work simply accumulated past states and information together in a driving behavior model to gain better prediction accuracy, although they also determined the optimal amount of past information required for ideal accuracy. There is limited work focused on the validation of the Markov property in the traffic realm. For example, \cite{shi2016research} conducted a comprehensive Markov analysis for driving cycle data and proved the existence of the Markov property for the evolution of macroscopic traffic flow indicators on urban networks, but such analysis is not focused on microscopic vehicle trajectories and driving behaviors. A significant gap in the literature exists due to the absence of rigorous analysis and empirical evidence confirming the applicability of the Markov property to driving behaviors. This gap persists due to the current lack of effective methodologies for quantitatively characterizing and detecting Markov properties in driving behaviors.

\par In the field of statistics, various kinds of statistical approaches have been established and applied to validate the existence of the Markov property in time series sequences. \cite{zhang2010markov} utilized a Chi-square-based statistical test method to prove the Markov property in the annual rainfall data. \cite{chen2010markov} developed a test method for the Markov property using the conditional characteristic function embedded in a frequency domain approach, and shown promising results in financial time series applications. \cite{shi2020does} proposed a Forward-Backward Learning procedure to test the Markov assumption in sequential decision-making, which is theoretically validated and empirically applied to a real dataset from mobile health studies. \cite{zhou2023testing} proposed a nonparametric test for the Markov property in high-dimensional time series using deep conditional generative learning. It is evident that there are currently numerous well-established statistical methods available for testing the Markov property in time series data. These methods have been widely applied in various fields, such as meteorology, healthcare, and finance. However, to date, no studies have been conducted to validate the applicability of these methods to driving trajectories and to employ similar approaches to examine whether the Markov property exists in driving behaviors from trajectory data.

\par In addition to validating whether driving behaviors adhere to the Markov property, comparing the differences in Markov property between AVs and HVs is also an important research gap that needs to be addressed. AVs rely on advanced algorithms and sensors to make decisions based on real-time data, which may lead to distinct patterns in their driving behaviors compared to human drivers. Human drivers, on the other hand, possess the ability to incorporate past knowledge, experiences, intuition, and anticipation into their decision-making process. Several studies in the cognitive realm have shown that the decision-making process for human drivers involves a complex perception, understanding, and information-processing process. The decisions made by human drivers are mostly based on the temporally accumulated information (\cite{mohammad2023cognitive,zgonnikov2024should, 10786483}). These findings reveal fundamental differences in the decision-making logic between AVs and HVs. Thus, examining whether the Markov property holds for both AVs and HVs can offer crucial insights into the fundamental mechanisms that govern their distinct driving behaviors.

\par In this study, we investigate the Markov property for driving behaviors with vehicle trajectories from a statistical testing perspective. We will utilize trajectories from both AVs and HVs obtained from public datasets. Our analysis will focus on examining the Markov property, determining the order of Markov, and calculating the percentage of trajectories exhibiting the Markov property for both AVs and HVs. Our findings ultimately demonstrate that HVs tend to exhibit weaker Markov properties compared to AVs. The percentage of Markov trajectories is lower for HVs than for AVs, and the order of Markov is higher for HVs compared to AVs. To the best of our knowledge, this study represents the first attempt to investigate the Markov property for vehicle trajectories from a statistical perspective. The results of this work will have significant implications for future related studies, as well as for the development of simulators and AV controllers.

\par The remainder of this paper is structured as follows: Section 2 provides a comprehensive description of the problem addressed in this study, including the definition of the MP and the specific car-following scenario considered. Section 3 presents a detailed explanation of the statistical methods employed in this research. Section 4 introduces the trajectory data utilized, including a description of the dataset and the data processing procedures. Section 5 reports the statistical results obtained from the analysis. Finally, Section 6 summarizes the entire study and discusses potential future research directions.

\section{Problem description}

\par In this section, we provide a fundamental description of the problem considered in this study. We begin by explaining the concept of the Markov property, which forms the basis of our analysis. Subsequently, the scenario utilized in validating the Markov property in driving behaviors is introduced. The notations employed in this section are shown in Table \ref{table1}.

\begin{table}[h!]
\renewcommand{\arraystretch}{1.3}
\caption{Notations in the problem description.}
\label{table1}
\centering
\begin{tabular}{|c|>{\centering\arraybackslash}p{14cm}|} \hline
Notation& Description \\ \hline  
\(t, T, \mathcal{T}\) & Time step, maximum time step, and the set of the time steps. \\ \hline  
\(v_{0, t}, v_{1, t}\)& Speed of the leading and following vehicle at time \(t\). \\ \hline  
\(a_{1, t}\)& Acceleration of the following vehicle at time \(t\). \\ \hline  
\(s_t\)& Distance between leading and following vehicle at time \(t\). \\ \hline  
\(A_t\)& Instant action of the following vehicle at time \(t\). \\ \hline  
\(\mathbf{X_t}\)& Instant state of the following vehicle at time \(t\). \\ \hline  
\(\mathbf{I_t}\)& History states set contains all past states of the following vehicle before time \(t\) (i.e. \(\mathbf{I_t} = \{\mathbf{X_t}, \mathbf{X_{t-1}}, \mathbf{X_{t-2}}, ...\}\)). \\ \hline 
\end{tabular}
\end{table}

\begin{figure*}[!t]
\centering
\includegraphics[width=1\textwidth]{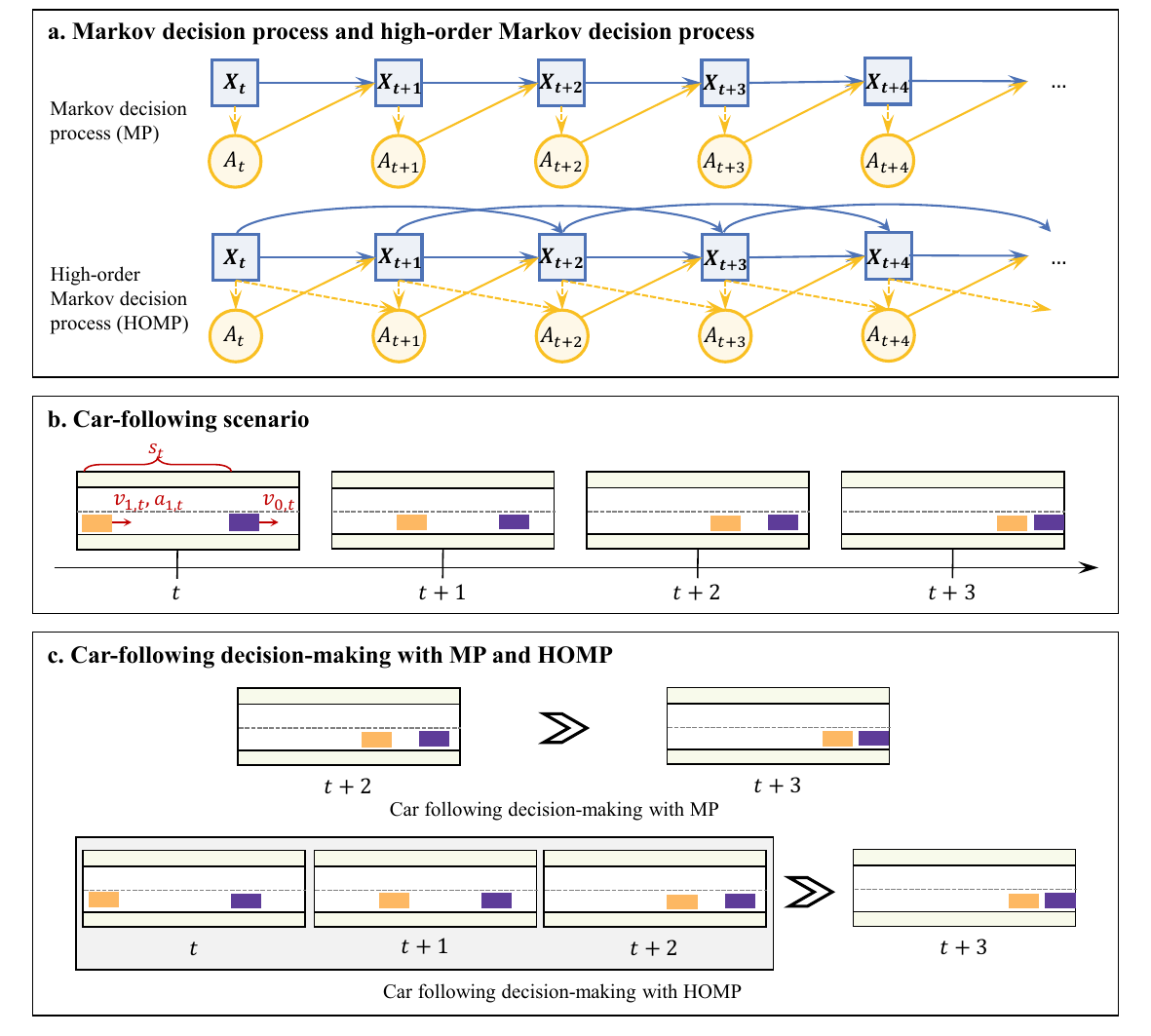}
\caption{Problem description. This study investigates the Markov property in car-following behavior by analyzing time series trajectories. We consider the leading vehicle's speed, the following vehicle's speed, and the distance between leading and following vehicle as state variables, and the following vehicle's acceleration as the decision variable. Our goal is to determine if the car-following driving process is a Markov process (MP) or a high-order Markov process (HOMP). If it is a HOMP, we also determine the order of the process based on the number of influential historical states to current action and future states.}
\label{problem desceiption}
\end{figure*}

\par As shown in Figure \ref{problem desceiption} (a), it is assumed in MP that the future state of the driving process depends only on its current state and not on past states. It is precise enough to predict the future only considering the current information. To be more specific, the driving action at time \(t\), \(A_t\), is made only considering the state at time \(t\), \(\mathbf{X_t}\). And the action \(A_t\) as well as the state \(\mathbf{X_t}\) directly determine the state at time \(t+1\), \(\mathbf{X_{t+1}}\). Such an assumption can be expressed as Equation \ref{MP}.

\begin{equation}
\label{MP}
\begin{aligned}
\mathbb{P}(\mathbf{X_{t+1}} \leq \mathbf{x}|\mathbf{I_t}) = \mathbb{P}(\mathbf{X_{t+1}} \leq \mathbf{x}|\mathbf{X_t}) 
\hbox{ for } \mathbf{x} \in \mathbb{R}^d, t\in \mathcal{T} \ \hbox{ and } \mathcal{T}=\{1, 2, ..., T\} 
\end{aligned}
\end{equation}

% suggestion：
% \begin{equation}
% \label{MP_revise}
% \begin{aligned}
% \mathbb{P}(\textbf{x}_{t+1} \leq \textbf{e}|I_t) = \mathbb{P}(\textbf{x}_{t+1} \leq \textbf{e}|\textbf{x}_t) \\
% \hbox{ for } \textbf{e} \in \mathbb{R}^d (or \textbf{e} \in \mathcal{X}) \hbox{ and } t\in \mathcal{T}\\
% \mathcal{T}:=\mathbb{N}
% \end{aligned}
% \end{equation}

As mentioned before, it may be difficult for MP to predict future states using only the current information. The High Order Markov decision Process (HOMP) can be employed to model more complex driving processes by extending the standard MP framework. In contrast to MP, HOMP allows the future state to be influenced by a finite number of past states (\cite{yu2023high}). To be more specific, in a HOMP, the driving action at time \(t\), \(A_{t}\), is made considering not only the current state \(\mathbf{X_t}\) but also its previous state \(\mathbf{X_{t-1}}, \mathbf{X_{t-2}}, ...\). Then the action \(A_{t}\) as well as the state \(\mathbf{X_t}\) determine the next state \(\mathbf{X_{t+1}}\). Since \(A_{t}\) is influenced by \(\mathbf{X_{t-1}}, \mathbf{X_{t-2}}, ...\), it can be defined that state \(\mathbf{X_{t+1}}\) will be determined not only by \(\mathbf{X_t}\), but also by the previous state \(\mathbf{X_{t-1}}, \mathbf{X_{t-2}}, ...\). Equation \ref{HOMP} defines the HOMP mathematically, in which the future state \(\mathbf{X_{t+1}}\) is determined by the current state \(\mathbf{X_t}\) and past state \(\{\mathbf{X_j}\}_{t-k<j < t}\), and it formulates a HOMP with a total of \(k-1\) past states considered. In Equation \ref{HOMP}, it is required that \(k\geq2\). As a special case, when \(k=1\), Equation \ref{HOMP} is the same as Equation \ref{MP}, which defines the MP.

\begin{equation}
\label{HOMP}
\begin{aligned}
   \mathbb{P}(\mathbf{X_{t+1}} \leq \mathbf{x}|I_t) = \mathbb{P}(\mathbf{X_{t+1}} \leq \mathbf{x}|\{\mathbf{X_j}\}_{t-k<j\leq t})  
   \hbox{ for } \mathbf{x} \in \mathbb{R}^d, k\geq2, t\in \mathcal{T} \hbox{ and } \mathcal{T}=\{1, 2, ..., T\}
\end{aligned}
\end{equation}

\begin{definition}
The Markov order represents the total number of current and past states that influence the state at the next time step. For a \(k\)-th order MP, when \(k=1\), it indicates that the state at the next moment is only influenced by the current state, which can be represented by Equation \ref{MP}. When \(k \geq 2\), it suggests that the state at the next moment is not only influenced by the current state but also related to the previous \(k-1\) historical states, as defined by Equation \ref{HOMP}.
\end{definition}

\begin{equation}
\label{define state}
\mathbf{X_t} = [v_{0, t}, v_{1, t}, s_{t}]^T
\hbox{ for } t\in \mathcal{T} \ \hbox{ and } \mathcal{T}=\{1, 2, ..., T\} 
\end{equation}

\begin{equation}
\label{define action}
A_t = a_{1, t}
\hbox{ for } t\in \mathcal{T} \ \hbox{ and } \mathcal{T}=\{1, 2, ..., T\} 
\end{equation}

\par In this study, we select the most basic car-following scenario to validate the Markov property in driving behaviors. As illustrated in Figure \ref{problem desceiption} (b), we choose time series trajectories recording car-following processes as samples for statistical testing. As defined in Equations \ref{define state} and \ref{define action},  in this study, the state variables include the speed of the leading vehicle \(v_{0,t}\), the speed of the following vehicle \(v_{1,t}\), and the distance between them \(s_{t}\), while the action variable is the acceleration of the following vehicle \(a_{1,t}\). Therefore, the state considered in this study is represented as a vector composed of three variables, i.e. \(d=3\) in Equations \ref{MP} and \ref{HOMP}. By verifying whether the decision at each moment and the state transition to the next moment are influenced only by the current state or jointly influenced by both the current and historical states, we aim to determine whether the car-following behavior possesses the Markov property. As depicted in Figure \ref{problem desceiption}(c), if the driving process adheres to the Markov property, we will classify it as an MP car-following process. Conversely, if the process does not exhibit the Markov property, we will regard it as a HOMP car-following process. In the latter case, we also determine the number of historical states that influence the current decision, thereby determining the order of the MP.

\section{Methodology}

\par This section first introduces the statistical test methods for verifying the Markov property of vehicle trajectories. Then, it presents the \(t\) test and \(F\) test methods used to determine whether there are significant differences in the Markov property between AV and HV trajectories.

\subsection{Markov Property Statistical Test}

\par This study aims to examine whether vehicle trajectories possess the Markov property. The Markov property statistical test method proposed by \cite{zhou2023testing} is utilized in this study. 

\par For any given vehicle trajectory, the null hypothesis for the statistical test is that the trajectory conforms to MP (as shown in Equation \ref{null hypothesis}), and the alternative hypothesis is that the trajectory does not possess the MP (Equation \ref{alternative hypothesis}, \cite{chen2012testing}).

\begin{equation}
\label{null hypothesis}
\begin{aligned}
    H_0: \mathbb{P}(\mathbf{X_{t+1}} \leq \mathbf{x}|\mathbf{I_t}) = \mathbb{P}(\mathbf{X_{t+1}} \leq \mathbf{x}|\mathbf{X_t}) 
    \hbox{for } \mathbf{x} \in \mathbb{R}^d,  t\in \mathcal{T} \hbox{ and } \mathcal{T}=\{1, 2, ..., T\}
\end{aligned}
\end{equation}

\begin{equation}
\label{alternative hypothesis}
\begin{aligned}
    H_A: \mathbb{P}(\mathbf{X_{t+1}} \leq \mathbf{x}|\mathbf{I_t}) \ne \mathbb{P}(\mathbf{X_{t+1}} \leq \mathbf{x}|\mathbf{X_t}) 
    \hbox{for } \mathbf{x} \in \mathbb{R}^d, t\in \mathcal{T} \hbox{ and } \mathcal{T}=\{1, 2, ..., T\}
\end{aligned}
\end{equation}

\par When the trajectory does not conform to the MP, it implies that we cannot simply use the current state to predict future states; instead, we need to consider some of the past states. We can verify whether the trajectory follows a HOMP by utilizing the null and alternative hypotheses shown in Equations \ref{HOMP null hypothesis} and \ref{HOMP alternative hypothesis}. These two hypotheses can be used to test whether the trajectory possesses \(k\)th order Markov property. Compared to Equation \ref{HOMP}, we expand the range of \(k\) from \(k \geq 2\) to  \(k \geq 1\), thus unifying the tests for both MP and HOMP. When \(k=1\), Equations \ref{HOMP null hypothesis} and \ref{HOMP alternative hypothesis} are identical to Equations \ref{null hypothesis} and \ref{alternative hypothesis}.

\begin{equation}
\label{HOMP null hypothesis}
\begin{aligned}
    H_0: \mathbb{P}(\mathbf{X_{t+1}} \leq \mathbf{x}|\mathbf{I_t}) = \mathbb{P}(\mathbf{X_{t+1}} \leq \mathbf{x}|\{\mathbf{X_j}\}_{t-k<j\leq t}) 
    \hbox{for } \mathbf{x} \in \mathbb{R}^d, k \geq 1, t\in \mathcal{T} \hbox{ and } \mathcal{T}=\{1, 2, ..., T\}
\end{aligned}
\end{equation}

\begin{equation}
\label{HOMP alternative hypothesis}
\begin{aligned}
    H_A: \mathbb{P}(\mathbf{X_{t+1}} \leq \mathbf{x}|\mathbf{I_t}) \ne \mathbb{P}(\mathbf{X_{t+1}} \leq \mathbf{x}|\{\mathbf{X_j}\}_{t-k<j\leq t}) 
    \hbox{for } \mathbf{x} \in \mathbb{R}^d, k \geq 1,   t\in \mathcal{T} \hbox{ and } \mathcal{T}=\{1, 2, ..., T\}
\end{aligned}
\end{equation}

\par Conditional Independence (CI) is commonly utilized in statistics to test the Markov property. The Markov property can be redefined using CI, stating that it holds under \(H_0\) if and only if the past and future states are independent, conditional on the present state. The notation shown in Equation \ref{CI} can be introduced to represent that the event \(\mathbf{Z_1}\) and \(\mathbf{Z_2}\) are independent, conditional on the event \(\mathbf{Z_3}\). To validate the hypothesis \(H_0\), it is necessary to conduct a series of tests for conditional independence shown in Equation \ref{Markov CI}.

\begin{equation}
\label{CI}
    \mathbf{Z_1} \perp \mathbf{Z_2} \mid \mathbf{Z_3}
\end{equation}

\begin{equation}
\label{Markov CI}
\begin{aligned}
    \mathbf{X_{t+k}} \perp \{\mathbf{X_j}\}_{t\leq j < t+k-1} \mid \mathbf{X_{t+k-1}} \hbox{ for } k \geq 1, t\in \mathcal{T} \hbox{ and } \mathcal{T}=\{1, 2, ..., T\}    
\end{aligned}
\end{equation}

\par The CI is usually characterized by Conditional Characteristic Function (CCF,  \cite{shi2020does}), which is the Fourier transform of the conditional probability density (\cite{chen2012testing}). For any vector \(\mathbf{Z_1}\), \(\mathbf{Z_2}\) and \(\mathbf{Z_3}\), Equation \ref{CI} will hold if and only if:
\begin{equation}
\label{CCF}
\begin{aligned}
    \mathbb{E}\{\exp(i \boldsymbol{\mu_1}^\top \mathbf{Z_1}) \mid \mathbf{Z_3}\}\mathbb{E}\{\exp(i \boldsymbol{\mu_2}^\top \mathbf{Z_2}) \mid \mathbf{Z_3}\} = 
    \mathbb{E}\{\exp(i \boldsymbol{\mu_1}^\top \mathbf{Z_1} + i \boldsymbol{\mu_2}^\top \mathbf{Z_2}) \mid \mathbf{Z_3}\} 
    \text{for }\boldsymbol{\mu_1} \in \mathbb{R}^d \text{ and  }\boldsymbol{\mu_2} \in \mathbb{R}^d     
\end{aligned}
\end{equation}

The CCF of \(\mathbf{X_t}\) given \(\mathbf{X_{t+1}}\) can be defined as Equation \ref{Markov CCF}  (\cite{zhou2023testing}). 

\begin{equation}
\begin{aligned}
\label{Markov CCF}
    \phi^*(\boldsymbol{\mu}|\mathbf{x}) = \mathbb{E}\{\exp(i\boldsymbol{\mu} ^ T \mathbf{X_{t+1}})|\mathbf{X_t} = \mathbf{x}\} 
    \text{for  }\boldsymbol{\mu} \in \mathbb{R}^d, t\in \mathcal{T} \hbox{ and } \mathcal{T}=\{1, 2, ..., T\}  
\end{aligned}
\end{equation}

\par Based on Equation \ref{Markov CCF}, the CI shown in Equation \ref{Markov CI} will hold if and only if the condition shown in Equation \ref{Markov CCF condition} exists.

\begin{equation}
\label{Markov CCF condition}
\begin{aligned}
    \phi^*(\boldsymbol{\mu}|\mathbf{X_{t+k-1}})\mathbb{E}[\exp(i\boldsymbol{\nu} ^ T \mathbf{X_t})|\{\mathbf{X_j}\}_{t<j<t+k}] 
    = \mathbb{E} [\exp(i\boldsymbol{\mu} ^ T \mathbf{X_{t+k}}+i\boldsymbol{\nu}^T\mathbf{X_t})|\{\mathbf{X_j}\}_{t<j<t+k}] \\
    \hbox{ for } k\geq 1, \boldsymbol{\mu}, \boldsymbol{\nu} \in \mathbb{R}^d, t\in \mathcal{T} \hbox{ and } \mathcal{T}=\{1, 2, ..., T\}  
\end{aligned}
\end{equation}

\par Taking another expectation on both sides of Equation \ref{Markov CCF condition}, it can be obtained:

\begin{equation}
\begin{aligned}
\label{Markov CCF condition 2}
    \mathbb{E}[\{\exp(i\boldsymbol{\mu} ^ T \mathbf{X_{t+k}}) - \phi ^* (\boldsymbol{\mu}|\mathbf{X_{t+k-1}})\}\exp(i\boldsymbol{\nu} ^ T \mathbf{X_t})]= 0  
    \hbox{ for } k\geq 1, \boldsymbol{\mu}, \boldsymbol{\nu} \in \mathbb{R}^d, t\in \mathcal{T} \hbox{ and } \mathcal{T}=\{1, 2, ..., T\}  
\end{aligned}
\end{equation}

\par The test statistic based on the Equation \ref{Markov CCF condition 2} is shown in Equation \ref{statistic markov old} (\cite{shi2020does, zhou2023testing}). 

\begin{equation}
\label {statistic markov old}
\begin{aligned}
    \tilde{S}(k, \boldsymbol{\mu}, \boldsymbol{\nu}) = 
    \frac{1}{T-k} \sum_{t=1}^{T-k} \left\{ \exp \left( i \boldsymbol{\mu}^\top \mathbf{X_{t+k}} \right) - \hat{\phi}(\boldsymbol{\mu} | \mathbf{X_{t+k-1}}) \right\} 
    \left\{ \exp \left( i \boldsymbol{\nu}^\top \mathbf{X_t} \right) - \bar{\phi}(\boldsymbol{\nu}) \right\}
\end{aligned}
\end{equation}

\par Calculating the test statistic shown in Equation \ref{statistic markov old} requires to estimate \(\phi ^*\) with estimator \(\hat{\phi}\). To accurately estimate \(\phi ^*\), \cite{zhou2023testing} proposed a doubly robust test statistic. The CCF of \(\mathbf{X_t}\) given \(\mathbf{X_{t+1}}\) is further defined as Equation \ref{Markov CCF new}.

\begin{equation}
\begin{aligned}
\label {Markov CCF new}
    \psi^*(\boldsymbol{\nu} \mid \mathbf{x}) = \mathbb{E}\left\{\exp\left(i\boldsymbol{\nu}^\top \mathbf{X_t}\right) \mid \mathbf{X_{t+1}} = \mathbf{x}\right\}  
    \text{for  }\boldsymbol{\nu} \in \mathbb{R}^d, t\in \mathcal{T} \hbox{ and } \mathcal{T}=\{1, 2, ..., T\}  
\end{aligned}
\end{equation}

\par Under \(H_0\), Equation \ref{Markov CCF condition 2} is further adapted into Equation \ref{Markov CCF condition new}.

\begin{equation}
\label {Markov CCF condition new}
\begin{aligned}
    \mathbb{E}\left\{\exp\left(i\boldsymbol{\mu}^\top \mathbf{X_{t+k}}\right) - \phi^*(\boldsymbol{\mu} \mid \mathbf{X_{t+k-1}})\right\} \times 
    \left\{\exp\left(i\boldsymbol{\nu}^\top \mathbf{X_t}\right) - \psi^*(\boldsymbol{\nu} \mid \mathbf{X_{t+1}})\right\} = 0. \\
    \hbox{ for } k\geq 1, \boldsymbol{\mu}, \boldsymbol{\nu} \in \mathbb{R}^d, t\in \mathcal{T} \hbox{ and } \mathcal{T}=\{1, 2, ..., T\}  
\end{aligned}
\end{equation}

\par The test statistic based on the Equation \ref{Markov CCF condition new} is shown as Equation \ref{statistic markov new}.

\begin{equation}
\label{statistic markov new}
\begin{aligned}
    S(k, \boldsymbol{\mu}, \boldsymbol{\nu}) = \frac{1}{T-k} \sum_{t=1}^{T-k} \left\{ \exp \left( i \boldsymbol{\mu}^\top \mathbf{X_{t+k}} \right) - 
    \hat{\phi}(\boldsymbol{\mu} | \mathbf{X_{t+k-1}}) \right\} 
    \left\{ \exp \left( i \boldsymbol{\nu}^\top \mathbf{X_t} \right) - \hat{\psi}(\boldsymbol{\nu} | \mathbf{X_{t+1}}) \right\}  
\end{aligned}
\end{equation}

\par \(\hat{\phi}, \hat{\psi}\) are the estimators of \(\phi^*\) and \(\psi^*\), the Mixture Density Network (MDN) was used to estimate them in the test procedures proposed in \cite{zhou2023testing}. A specific significance level \(\alpha\), is chosen to choose to accept or reject the null hypothesis. In \cite{zhou2023testing}, a significance level of 0.05 was used. In this study, we adopte the significance levels \(\alpha\) of 0.01 and 0.05 as thresholds to accept or reject the null hypothesis. When the \(p\) value is greater than the significance level \(\alpha\), the null hypothesis is accepted, indicating the presence of the Markov property; otherwise, the null hypothesis is rejected.

\par To determine the Markov order \(k\) of the trajectory, the test will be conducted multiple times with different values of \(k\). Suppose the trajectory follows a \( K \)-th order Markov model. Then the null hypothesis \( H_0 \) holds for  \( k \geq K \), but does not hold for any \( k < K \). We set the estimated order to be the first integer \( k \) for which we fail to reject \( H_0 \).

% \hrule
% \vspace{0.2cm}
% \noindent\textbf{Algorithm: Determining Markov order} \\
% \hrule
% \vspace{0.2cm}

% \noindent\textbf{Input:} $k$ and the trajectory data. \\
% \textbf{for} $k = 1, 2, \dots, K$ \textbf{do} \\
% \hspace*{1em} Conduct Markov property test.\\
% \hspace*{1em} \textbf{if} $H_0$ is not rejected \textbf{then} \\
% \hspace*{2em} Conclude the trajectory is from a $k$-th order HOMP; \textbf{Break.} \\
% \hspace*{1em} \textbf{end if} \\
% \textbf{end for} \\

% \hrule

\subsection{t Statistical Test}
\par Through the statistical test of the Markov property, we can obtain the corresponding Markov order for each car-following trajectory. To draw general conclusions about the difference in Markov property of AV and HV car-following trajectories, we will conduct the Markov statistical tests on multiple trajectories to obtain the respective Markov order for each type of trajectory. To describe whether there is a significant difference between the Markov order of AV and HV trajectories, we use the two-sample \( t\) test method for testing.

\par The test statistic \(t\) is calculated using Equations \ref{test statistic t} - \ref{degrees of freedom for t test}. \(\bar{K}_{HV}\) and \(\bar{K}_{AV}\) represent the Markov order means for HV trajectories and AV trajectories, respectively. \(n_{HV}\) and \(n_{AV}\) denote the number of HV and AV trajectories. \(K_{HV, i}\) and \(K_{AV, i}\) refer to the \(i\)th HV and AV trajectory's Markov order. \(S_{HV}^2\) and \(S_{AV}^2\) are the variances of the HV and AV trajectories' Markov order, respectively. \(S_p\) is the pooled standard deviation. \(d_t\) is the degrees of freedom in the \(t\) test. Equation \ref{test statistic t} defines the calculation for the test statistic \(t\). Equations \ref{order mean hv} and \ref{order mean av} define the calculations for the Markov order means \(\bar{K}_{HV}\) and \(\bar{K}_{AV}\), respectively. Equations \ref{order variance hv}, \ref{order variance av}, and \ref{pooled standard deviation} define the calculations for the Markov order variances and the pooled standard deviation. Equation \ref{degrees of freedom for t test} shows the calculation for the degrees of freedom of the \(t\) test.

\begin{equation}
    \label{test statistic t}
    t = \frac{\bar{K}_{HV} - \bar{K}_{AV}}{S_p \sqrt{\frac{1}{n_{HV}} + \frac{1}{n_{AV}}}}
\end{equation}

\begin{equation}
    \label{order mean hv}
    \bar{K}_{HV} = \frac{1}{n_{HV}} \sum_{i=1}^{n_{HV}} K_{HV,i}
\end{equation}

\begin{equation}
    \label{order mean av}
    \bar{K}_{AV} = \frac{1}{n_{AV}} \sum_{i=1}^{n_{AV}} K_{AV,i}
\end{equation}

\begin{equation}
    \label{order variance hv}
    S_{HV}^{2} = \frac{1}{n_{HV} - 1} \sum_{i=1}^{n_{HV}} (K_{HV,i} - \bar{K}_{HV})^2
\end{equation}

\begin{equation}
    \label{order variance av}
    S_{AV}^{2} = \frac{1}{n_{AV} - 1} \sum_{i=1}^{n_{AV}} (K_{AV,i} - \bar{K}_{AV})^2
\end{equation}

\begin{equation}
    \label{pooled standard deviation}
    S_p = \sqrt{\frac{(n_{HV} - 1)S_{HV}^2 + (n_{AV} - 1)S_{AV}^2}{n_{HV} + n_{AV} - 2}}
\end{equation}

\begin{equation}
    \label{degrees of freedom for t test}
    d_t = n_{HV} + n_{AV} - 2
\end{equation}

\par The \(p\) value for the \(t\) test is calculated using the Cumulative Distribution Function (CDF) of the \(t\) distribution. Since we expect to prove whether the Markov order for HV trajectories is significantly larger than that of AV trajectories, the one-tailed test is utilized. The \(p\) value for the one-tailed test is shown in Equation \ref{p value for t test}.

\begin{equation}
    \label{p value for t test}
    p = 1 - \text{CDF}_t(t, d_t)
\end{equation}

\par If the test statistic \(t\) is larger than 0, and \(\frac{p}{2}\) is smaller than the threshold 0.05, it will be concluded that the mean of the Markov order for HV trajectories is significantly greater than that of AV trajectories.

\subsection{F Statistical Test}

\par To compare whether there is a significant difference in the variance of the Markov order distribution between HV and AV trajectories, that is, to determine if HV and AV exhibit heterogeneous differences in their car-following decision-making processes, we used the \(F\) test to compare the variances of the Markov order distributions for HV and AV.

\par The test statistic \(F\) is calculated using Equations \ref{test statistic f} - \ref{degree 2 of freedom for f test} and Equations \ref{order mean hv}  - \ref{order variance av}. \(d_{f, HV}\) and \(d_{f, AV}\) are the degree of freedom in the \(F\) test. Equation \ref{test statistic f} defines the calculation for the test statistic \(F\) based on the Markov order variance \(S_{HV}^{2}\) and \(S_{AV}^{2}\). Equations \ref{degree 1 of freedom for f test} and \ref{degree 2 of freedom for f test} show the calculation for the degrees of freedom of the \(F\) test.

\begin{equation}
    \label{test statistic f}
    F = \frac{S_{HV}^2}{S_{AV}^2}
\end{equation}

\begin{equation}
    \label{degree 1 of freedom for f test}
    d_{f, HV} = n_{HV} - 1
\end{equation}

\begin{equation}
    \label{degree 2 of freedom for f test}
    d_{f, AV} = n_{AV} - 1
\end{equation}

\par The \(p\) value for the \(F\) test is calculated using the CDF of the \(F\) distribution. The \(p\) value for the \(F\) test is shown in Equation \ref{p value for f test}.

\begin{equation}
    \label{p value for f test}
    p = 1 - \text{CDF}_F(F, d_{f, HV}, d_{f, AV})
\end{equation}

\par If the value of \(p\) is greater than the threshold 0.95, it will be concluded that the variance of the Markov order for HV trajectories is significantly greater than that of AV trajectories. HVs exhibit larger heterogeneity than AVs in their car-following decision-making processes.

\section{Trajectory data}
\par Trajectory datasets used in this study and the data processing details are given in this section. A summary of the utilized trajectories, including their quantity and length, is provided.

\subsection{Lyft Level-5 Dataset}
\par The Lyft Level 5 dataset was collected by a fleet of self-driving vehicles equipped with 7 cameras, 3 LiDARs (one 64-channel roof-mounted spinning at 10 Hz and two 40-channel front-bumper-mounted), and 5 radars (four roof-mounted and one forward-facing front-bumper-mounted), providing a 360° horizontal field of view (\cite{houston2021one}). The data was captured during the daytime (8 AM to 4 PM) between October 2019 and March 2020 along a fixed route. Visible traffic participants, including vehicles, pedestrians, and cyclists, were detected using an in-house perception system. Each traffic participant was represented by a 2.5D cuboid, velocity, acceleration, yaw, yaw rate, and a class label, with the perception system fusing data across multiple modalities to produce a 360° view of the surrounding world.
\par \cite{li2023large} processed and enhanced car-following trajectories from the Lyft Level-5 data focusing on various categories: HV-following-AV, HV-following-HV, and AV-following-HV. The process involved selecting car-following pairs based on specific rules, assessing the quality of raw data through anomaly analysis, and then correcting and enhancing the raw car-following data using motion planning, Kalman filtering, and wavelet denoising techniques. The processed dataset contains more than 70,000 car-following trajectories, covering a total driving distance of over 150,000 kilometers. This study utilized these processed car-following trajectories for Markov property analysis.
\subsection{CATS Lab ACC Dataset}
\par The CATS Lab ACC dataset was collected and publicly shared by the Connected and Autonomous Transportation Systems Laboratory (CATS Lab, \cite{shi2021empirical}). The dataset consists of trajectories from five vehicles (two AVs: Lincoln MKZ 2016 and 2017; three HVs) arranged from downstream to upstream as HV, AV, AV, HV, and HV. Each vehicle was equipped with a uBox C066-F9P GPS device (location accuracy: 0.26 m; speed accuracy: 0.089 m/s; data frequency: 10 Hz). Data were collected at two locations on open public roads with surrounding traffic disturbances. Both tests were conducted on clear nights with street lighting and vehicle headlights, and overtaking was prohibited.

\par The first low-speed dataset in CATS Lab ACC dataset was collected on a 2.4 km segment of Lizard Trail Road, Tampa, Florida, USA, on November 18th, 2020, with a maximum speed of about 15 m/s. The second high-speed dataset was collected on an 8 km segment of State Road 56, Tampa, Florida, USA, on November 24th, 2020, with a maximum speed of roughly 29 m/s. During the experiments, the first vehicle was instructed to generate varied oscillation patterns by regularly accelerating and decelerating, while AVs followed preceding vehicles with ACC turned on, and HVs followed as usual.

\par The car-following trajectories for both AVs and HVs are adopted in this study to validate the Markov property.

\subsection{Data Processing}
\par The following processing procedures were applied to two trajectory datasets in this study:
\begin{enumerate}
    \item Vehicle locations were calculated based on longitude and latitude in the CATS Lab ACC dataset.
    \item Linear interpolation was used to fill in missing GPS information in the CATS Lab ACC dataset.
    \item The data was resampled to 10 Hz, resulting in a 1-second interval between consecutive data points.
    \item State variables and action variables were calculated from vehicle locations using one-order differential and kinematic equations.
    \item To ensure consistency in the length of trajectories used for the statistical test, we divided the long car-following trajectory in the CATS Lab ACC dataset into equal-length segments of 120 seconds. In the Lyft Level-5 dataset processed by \cite{li2023large}, we retained trajectories with a length greater than 70 seconds.
    \item Data from the beginning or end of the test runs in the CATS Lab ACC dataset was excluded.
\end{enumerate}

\begin{table}[!h]
\caption{Summary of car-following trajectories in Lyft level-5 dataset and CATS lab ACC dataset to examine the Markov property.}
\label{trajectory dataset info}
\begin{tabular}{|>{\centering\arraybackslash}p{1.5cm}|>{\centering\arraybackslash}p{3.5cm}|>{\centering\arraybackslash}p{2cm}|>{\centering\arraybackslash}p{2.5cm}|>{\centering\arraybackslash}p{3.5cm}|} \hline  
Dataset& Trajectory type& Trajectory quantity& Average length (s)& Standard deviation of length (s) \\ \hline  
\multirow{2}{4em}{Lyft Level-5} &AV&21&77.50 & 9.64\\ \cline{2-5}
~ &HV&134&96.84 & 32.66\\ \hline 
\multirow{6}{3em}{CATS Lab ACC} &AV Cruising (AV-C)& 15 &120.00 &0.00 \\ \cline{2-5}
~ &AV Oscillation (AV-O)& 39 & 120.00&0.00 \\ \cline{2-5}
~ &AV all (AV)&54 & 120.00&0.00 \\ \cline{2-5}
~ &HV Cruising  (HV-C)& 20 & 120.00&0.00 \\ \cline{2-5}
~ &HV Oscillation (HV-O)& 41& 120.00&0.00 \\ \cline{2-5}
~ &HV all (HV)&61 & 120.00&0.00 \\ \hline 
\end{tabular}
\end{table}

\par The processed trajectory data information is shown in Table \ref{trajectory dataset info}. We divided the Lyft Level-5 dataset into two categories: trajectories where the following vehicle is an AV and trajectories where the following vehicle is an HV, to explore the differences in Markov property exhibited by AVs and HVs when making car-following decisions. We obtained a total of 21 AV car-following trajectories and 134 HV car-following trajectories, with average lengths of 77.50s and 96.84s, respectively. The longest AV car-following trajectory is 87.14s, and the longest HV trajectory is 129.50s.

\par Since the car-following experiments in CATS Lab ACC dataset were set up with precise cruising and oscillation scenarios, we divided trajectories in the CATS Lab ACC dataset into four categories: AV trajectories during Cruising (AV-C) and Oscillation (AV-O), and HV trajectories during Cruising (HV-C) and Oscillation (HV-O), to study the differences in Markov property among them. There are 15 AV-C trajectories, 39 AV-O trajectories, 20 HV-C trajectories, and 41 HV-O trajectories. In total, we obtained 54 AV and 61 HV car-following trajectories. All trajectories have a length of 120s.

\section{Results}

\par This section discusses the outcomes of the Markov property statistical tests conducted on car-following trajectories from two datasets. The results include an analysis of the Markov property, the determined order of the Markov process, and the proportion of trajectories displaying Markov property for both HVs and AVs. The findings are presented separately for each dataset.

\subsection{Lyft Level-5 Dataset}

\par
\begin{figure}[!t]
\centering %图片居中
\includegraphics[width=0.5\textwidth]{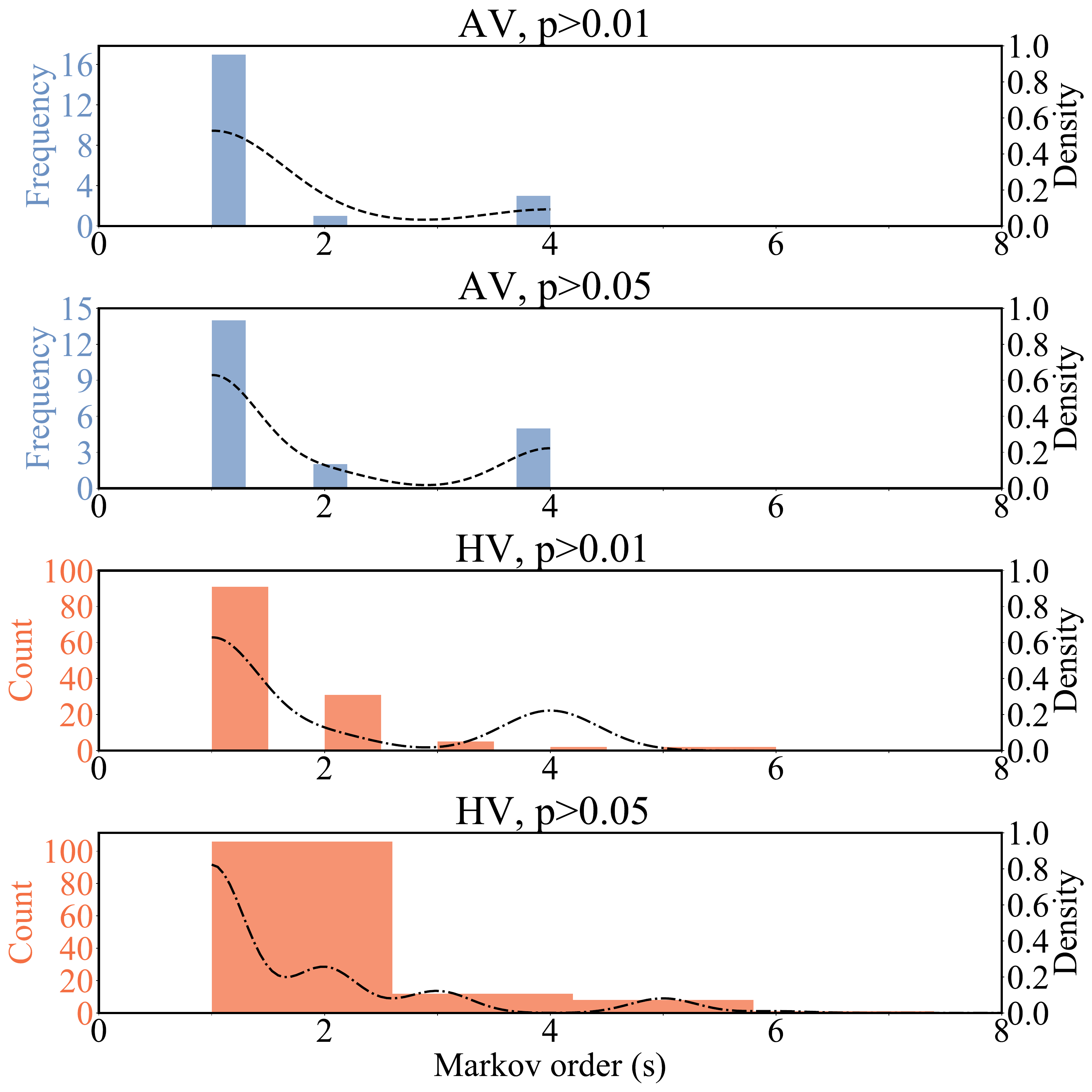}
\caption{Distribution of Markov orders for car-following trajectories in the Lyft Level-5 dataset. The first two plots (blue bars) show the Markov orders for AVs, and the last two plots (orange bars) show the Markov orders for HVs. The statistical test results with both \(p>0.01\) and \(p>0.05\) are presented.}
\label{Fig_lyft_hist}
\end{figure}

\begin{figure}[!t]
\centering
\includegraphics[width=0.5\textwidth]{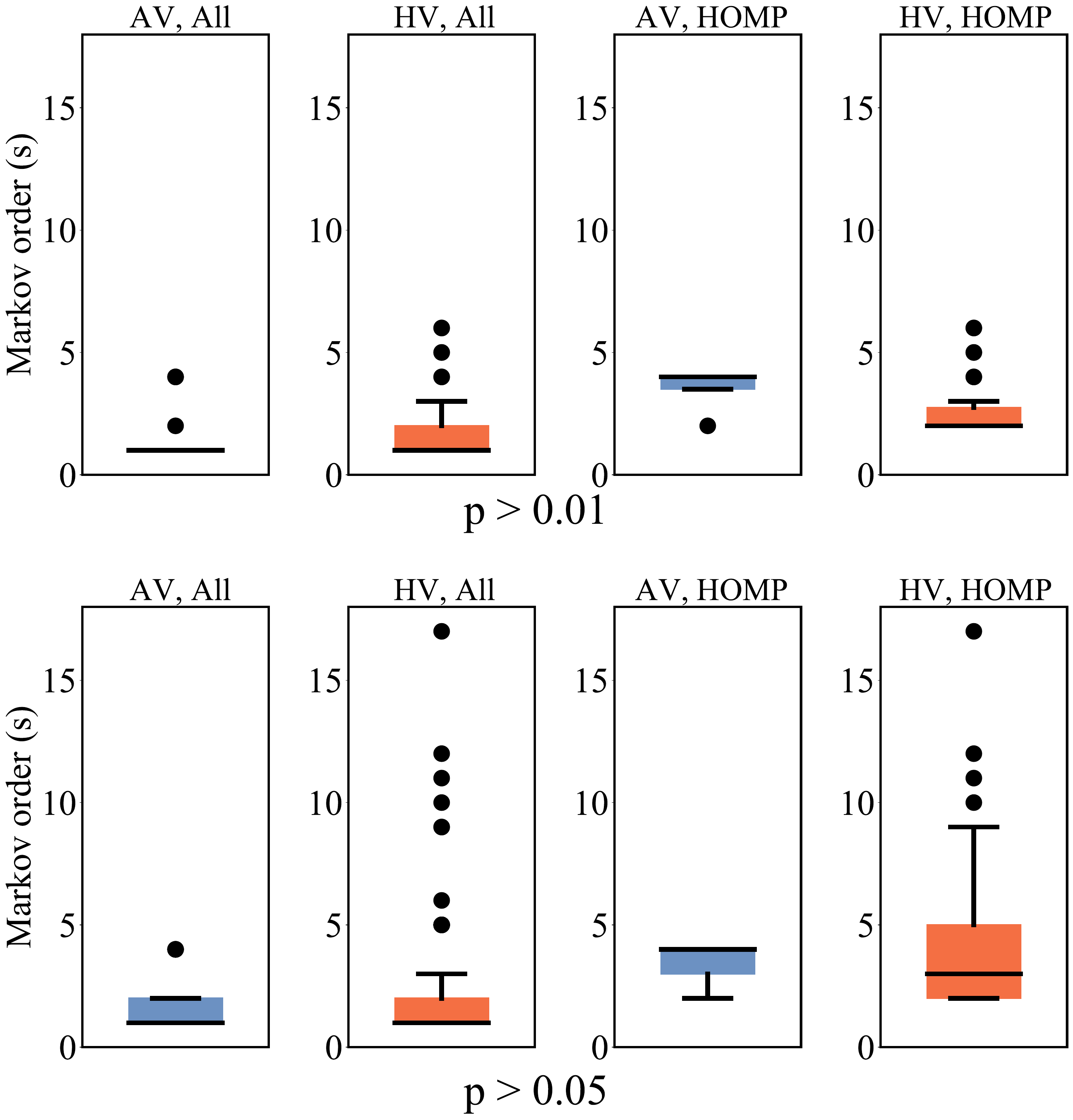}
\caption{Box plots of Markov orders for car-following trajectories in the Lyft Level-5 dataset. The first and third columns, with blue boxes, represent the Markov orders of AVs. The second and fourth columns, with orange boxes, represent the Markov orders of HVs. The first two columns show the orders for the combined trajectory set of MP and HOMP. The last two columns show the orders for HOMP trajectories only. The first row shows the results for \(p > 0.01\), and the second row shows the results for \(p > 0.05\).}
\label{Fig_lyft_box}
\end{figure}

 We conducted statistical tests on all 21 AV trajectories and 134 HV trajectories in Lyft Level-5 dataset listed in Table \ref{trajectory dataset info}, recording the \(p\) values for each trajectory when the Markov order ranges from 1 to 10, as well as whether the null hypothesis holds. We selected the smallest Markov order that fails to reject the null hypothesis to hold as the Markov order of the trajectory. 
The Markov property verification results for car-following trajectories in the Lyft Level-5 dataset are shown in Table \ref{Lyft p_0.01}, Table \ref{Lyft p_0.05}, Table \ref{Lyft difference}, Figure \ref{Fig_lyft_hist}, and Figure \ref{Fig_lyft_box}.

\begin{table}[!h]
\caption{Summary of the Markov property statistical test results (\(p>0.01\)) for trajectories in Lyft level-5 dataset.}
\label{Lyft p_0.01}
\begin{tabular}{|>{\centering\arraybackslash}m{2cm}|>{\centering\arraybackslash}m{3cm}|>{\centering\arraybackslash}m{3cm}|>{\centering\arraybackslash}m{3cm}|>{\centering\arraybackslash}m{3cm}|} \hline
Trajectory set& Mean of the Markov order (s)& Standard deviation of Markov order (s)& Percentage of MP trajectories (\%)& Percentage of HOMP trajectories (\%) \\ \hline
AV &1.48&1.08& 80.95& 19.05\\ \hline
HV&1.49&0.95& 67.16&32.84\\ \hline
\end{tabular}
\end{table}

\begin{table}[!h]
\caption{Summary of the Markov property statistical test results (\(p>0.01\)) for trajectories in Lyft level-5 dataset.}
\label{Lyft p_0.05}
\begin{tabular}{|>{\centering\arraybackslash}m{2cm}|>{\centering\arraybackslash}m{3cm}|>{\centering\arraybackslash}m{3cm}|>{\centering\arraybackslash}m{3cm}|>{\centering\arraybackslash}m{3cm}|} \hline
Trajectory set& Mean of the Markov order (s)& Standard deviation of Markov order (s)& Percentage of MP trajectories (\%)& Percentage of HOMP trajectories (\%) \\ \hline
AV &1.81&1.29& 71.43& 28.57\\ \hline
HV&2.11&2.38& 60.44&39.56\\ \hline
\end{tabular}
\end{table}

\begin{table}[!h]
\caption{Statistical test for the difference in Markov order between AV and HV trajectories in Lyft level-5 dataset.}
\label{Lyft difference}
\begin{tabular}{|>{\centering\arraybackslash}m{2cm}|>{\centering\arraybackslash}m{2cm}|>{\centering\arraybackslash}m{2cm}|>{\centering\arraybackslash}m{2cm}|>{\centering\arraybackslash}m{2cm}|>{\centering\arraybackslash}m{2cm}|} \hline
Trajectory sets& \(p\) value for Markov test & \(t\) statistic & \(p\) value for \(t\) statistic & \(F\) statistic & \(p\) value for \(F\) statistic \\ \hline
AV / HV & 0.1 & 0.0551 &  0.9561 & 0.7774 & 0.1990 \\ \hline
AV / HV & 0.5 & 0.5699 &  0.5695 & \textbf{3.4065} & \textbf{0.9987} \\ \hline
\end{tabular}
\end{table}

\par Table \ref{Lyft p_0.01} and Table \ref{Lyft p_0.05} present the Markov statistical test results with the threshold of \(p>0.01\) and \(p>0.05\), respectively. Comparatively, \(p>0.01\) is an easier threshold to meet, while \(p>0.05\) is a more challenging standard. The tables record the average Markov order of all trajectories, the standard deviation of the Markov order for all trajectories, and the proportion of trajectories that conform to MP (Markov order of 1 second) and the proportion of trajectories that conform to HOMP (Markov order greater than 1 second) among all tested trajectories.

\par Table \ref{Lyft difference} records the differences in Markov order exhibited by the tested AV and HV car-following trajectories in Lyft Level-5 dataset. The \(t\) test is used to determine whether the Markov order of the HV trajectory is significantly higher than that of the AV trajectory. When the value of the \(t\) statistic is greater than 0 and the value of \(\frac{p}{2}\) is less than 0.05, it is considered that the Markov order of the HV trajectory is significantly higher than that of the AV trajectory. The \(F\) test is employed to examine whether the variance of the Markov order of the HV trajectory is significantly higher than that of AV. When the \(p\) value is greater than 0.95, it is deemed that the variance of the Markov order of the HV trajectory is significantly higher than that of AV.

\par When the null hypothesis of the Markov property test is accepted for the \(p\) value greater than 0.01, the average Markov order for the AV trajectory is 1.48 seconds, and for the HV trajectory, it is 1.49 seconds. The \(t\) test results shown in Table \ref{Lyft difference} indicate that the Markov order of HV is not significantly higher than that of AV (\(t\) statistic value is 0.0551, and the \(p\) value is 0.9561). The \(F\) test results show that the variance of the Markov order of HV is also not significantly higher than that of AV (\(F\) statistic value is 0.7774, and the \(p\) value is 0.1990). However, among the AV trajectories, 80.95\% conform to the MP, and 19.05\% conform to the HOMP. In contrast, among the HV trajectories, only 67.16\% conform to the MP, and 32.84\% conform to HOMP. Therefore, it can be concluded that when the threshold for the null hypothesis of the Markov property test to hold is the \(p\) value greater than 0.01, the average and variance of the Markov order for AV and HV in the Lyft Level-5 dataset do not differ significantly, but the proportion of AV trajectories conforming to MP is higher than that of HV.

\par If the null hypothesis of the Markov property test is accepted for the \(p\) value greater than 0.05, the average Markov order for AV is 1.81 seconds, and for HV, it is 2.11 seconds, showing no obvious difference between AV and HV. The standard deviation of the Markov order for AV is 1.29 seconds, and for HV, it is 2.38 seconds, with AV's standard deviation being smaller than HV's. The \(t\) test and \(F\) test results in Table \ref{Lyft difference} also support this conclusion. The \(t\) statistic value is 0.5699, and the \(p\) value is 0.5695, indicating that there is no significant difference in the Markov order between HV and AV trajectories. The \(F\) statistic value is 3.4065, and the \(p\) value is 0.9987, suggesting that the variance of the Markov order for HV trajectories is significantly higher than that of AV. Besides, among the AV trajectories, 71.43\% conform to the MP, while only 60.44\% of the HV trajectories conform to MP. Consequently, it can be concluded that when the threshold for the null hypothesis to hold is the \(p\) value greater than 0.05, in the Lyft Level-5 dataset, the average Markov order of AV is similar to that of HV, but the distribution of AV's Markov order is more concentrated compared to HV, and the proportion of AV trajectories conforming to MP is higher than that of HV.

\par Figure \ref{Fig_lyft_hist}, and Figure \ref{Fig_lyft_box} visualize the Markov order of all trajectories in the Lyft Level-5 dataset. Figure \ref{Fig_lyft_hist} shows the frequency distribution and probability density curve of the Markov order, while Figure \ref{Fig_lyft_box} presents box plots of the Markov order distribution. The results for both thresholds for the Markov property statistical test, \(p>0.01\) and \(p>0.05\), are presented. From these two figures, it can be observed that when the threshold for the Markov property to hold is \(p>0.01\), there is no significant difference in the distribution of the Markov order between AV and HV trajectories. When the threshold for the Markov property to hold is \(p>0.05\), the mean values of the Markov order for AV and HV trajectories do not differ greatly. However, the distribution of the Markov order for HV trajectories is noticeably wider than that of AV: in Figure \ref{Fig_lyft_box}, the orange box plot has more outliers compared to the blue box plot. These outliers indicate that among the HV trajectories represented by the orange box, some trajectories exhibit high-order Markov properties, suggesting the presence of longer memory in these car-following trajectories. The decision-making process in these trajectories considers a longer period of historical information.

\par To summarize, in the Lyft Level-5 dataset, we selected a total of 21 AV car-following trajectories and 134 HV car-following trajectories. Based on all the results, the following conclusions can be drawn. When choosing \(p>0.05\) as the threshold for the null hypothesis to hold in the Markov property statistical test, the average Markov order of AV trajectories is 1.81s, while the average Markov order of HV trajectories is 2.11s. The results of the \(t\) test indicate that the Markov order of HV trajectories is similar to that of AV trajectories. The results of the \(F\) test show that the distribution of the Markov order for HV trajectories is noticeably wider than that of AV trajectories. The Markov order of AV trajectories is mostly concentrated at lower values, suggesting that AVs are more likely to consider only the current real-time state information when making car-following decisions. In contrast, HV trajectories have several high-order outliers in their Markov order, indicating that HVs consider more historical information than AVs when making car-following decisions. Additionally, the length of historical information considered in the decision-making process varies among different HV trajectories, with some HV trajectories considering only short-term historical information while others consider long-term historical information. The decision-making process of HV exhibits heterogeneity.

\subsection{CATS Lab ACC Dataset}

\par
\begin{figure*}[!t]
\centering %图片居中
\includegraphics[width=0.9\textwidth]{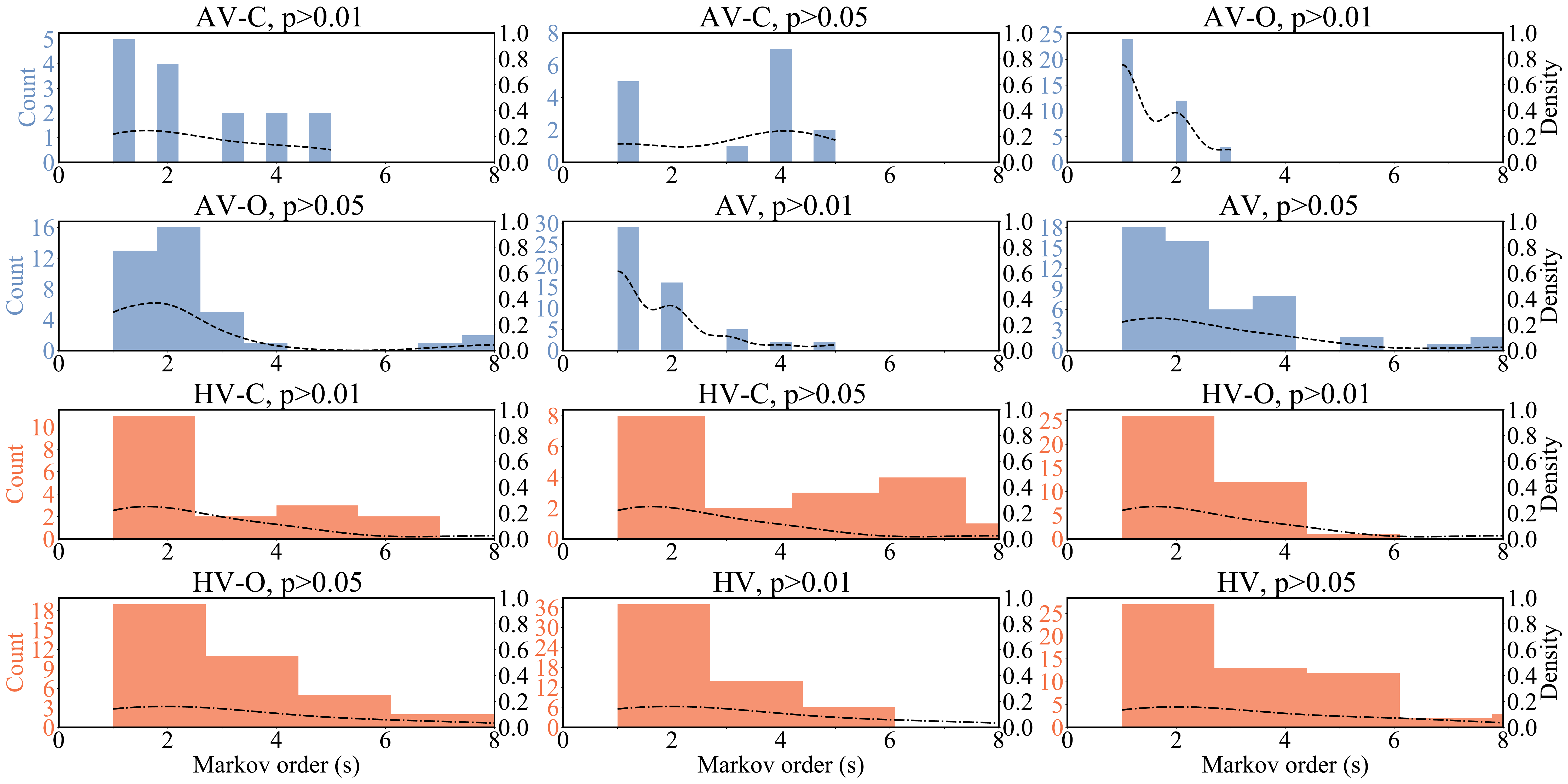}
\caption{Distribution of Markov orders for car-following trajectories in the CATS lab ACC dataset. The first two rows (blue bars) of the distribution plots show the Markov orders for AVs, and the last two rows (orange bars) show the Markov orders for HVs. The statistical test results with both \(p>0.01\) and \(p>0.05\) are presented.}
\label{Fig_catsacc_hist}
\end{figure*}

\begin{figure*}[!t]
\centering
\includegraphics[width=0.9\textwidth]{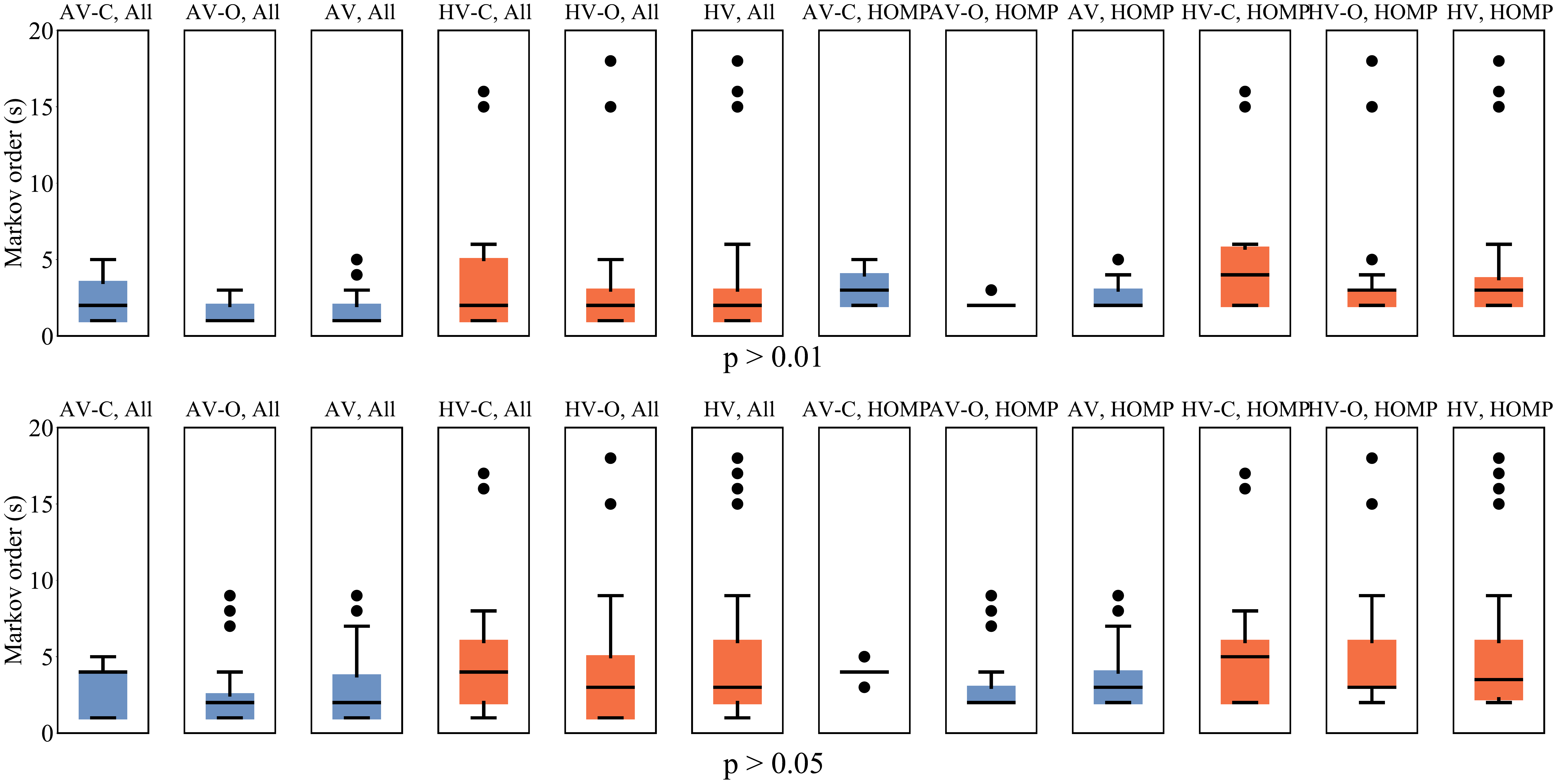}
\caption{Box plot of Markov orders for car-following trajectories in the CATS lab ACC dataset. The columns with blue boxes represent the Markov orders of AVs. The other columns with orange boxes, represent the Markov orders of HVs. The first six columns show the orders for the combined trajectory set of MP and HOMP. The last six columns show the orders for HOMP trajectories only. The first row shows the results for \(p > 0.01\), and the second row shows the results for \(p > 0.05\).}
\label{Fig_catsacc_box}
\end{figure*}

 We then conducted statistical tests on all 54 AV trajectories and 61 HV trajectories in the CATS Lab ACC Dataset listed in Table \ref{trajectory dataset info}, recording the \(p\) values for each trajectory when the Markov order ranges from 1 to 10, as well as whether the null hypothesis holds. The Markov statistical results for car-following trajectories in the CATS Lab ACC dataset are shown in Table \ref{CATS p_0.01}, Table \ref{CATS p_0.05}, Table \ref{cats difference}, Figure \ref{Fig_catsacc_hist} and Figure \ref{Fig_catsacc_box}. The explanations of the organization of Table \ref{CATS p_0.01}, Table \ref{CATS p_0.05}, Table \ref{cats difference}, and details of \(t\) test and \(F\) test are similar to Table \ref{Lyft p_0.01}, Table \ref{Lyft p_0.05}, Table \ref{Lyft difference}.

\begin{table}[!h]
\caption{Summary of the Markov property statistical test results (\(p>0.01\)) for trajectories in CATS lab ACC dataset.}
\label{CATS p_0.01}
\begin{tabular}{|>{\centering\arraybackslash}m{2cm}|>{\centering\arraybackslash}m{3cm}|>{\centering\arraybackslash}m{3cm}|>{\centering\arraybackslash}m{3cm}|>{\centering\arraybackslash}m{3cm}|} \hline
Trajectory set& Mean of the Markov order (s)& Standard deviation of Markov order (s)& Percentage of MP trajectories (\%)& Percentage of HOMP trajectories (\%) \\ \hline
AV-C&2.47&1.46& 33.33& 66.67\\ \hline
AV-O&1.47&0.64& 61.53& 38.47\\ \hline
AV&1.74 & 1.03&53.70 & 46.30\\ \hline
HV-C&4.00&4.30&30.00&70.00\\ \hline
HV-O&2.78&3.31&31.71&68.29\\ \hline
HV&3.18 & 3.67&31.15 &68.85 \\ \hline
\end{tabular}
\end{table}

\begin{table}[!h]
\caption{Summary of the Markov property statistical test results (\(p>0.05\)) for trajectories in CATS lab ACC dataset.}
\label{CATS p_0.05}
\begin{tabular}{|>{\centering\arraybackslash}m{2cm}|>{\centering\arraybackslash}m{3cm}|>{\centering\arraybackslash}m{3cm}|>{\centering\arraybackslash}m{3cm}|>{\centering\arraybackslash}m{3cm}|} \hline
Trajectory set& Mean of the Markov order (s)& Standard deviation of Markov order (s)& Percentage of MP trajectories (\%)& Percentage of HOMP trajectories (\%) \\ \hline
AV-C&3.07&1.58& 33.33& 66.67\\ \hline
AV-O&2.46&2.05& 33.33& 66.67\\ \hline
AV&2.63 &1.94 &33.33 & 66.67\\ \hline
HV-C&4.95&4.54&10.00&90.00\\ \hline
HV-O&3.71&3.66&31.71&68.29\\ \hline
HV&4.13 &3.94 &24.59 & 75.41\\ \hline
\end{tabular}
\end{table}

\begin{table}[!h]
\caption{Statistical test for the difference in Markov order between AV and HV trajectories in CATS Lab ACC dataset.}
\label{cats difference}
\begin{tabular}{|>{\centering\arraybackslash}m{2cm}|>{\centering\arraybackslash}m{2cm}|>{\centering\arraybackslash}m{2cm}|>{\centering\arraybackslash}m{2cm}|>{\centering\arraybackslash}m{2cm}|>{\centering\arraybackslash}m{2cm}|} \hline
Trajectory sets& \(p\) value for Markov test & \(t\) statistic & \(p\) value for \(t\) statistic & \(F\) statistic & \(p\) value for \(F\) statistic \\ \hline
AV-C / HV-C & 0.1 & 1.3198 & 0.1959 & \textbf{8.7231} & \textbf{0.9999} \\ \hline
AV-C / HV-C & 0.5 & 1.6118 &  0.1165 & \textbf{7.8465} & \textbf{0.9998} \\ 
\hline
AV-O / HV-O & 0.1 & \textbf{2.4476} &  \textbf{0.0166} & \textbf{26.4571} & \textbf{0.9999} \\ \hline
AV-O / HV-O & 0.5 & \textbf{1.8642} &  \textbf{0.0660} & \textbf{3.1915} & \textbf{0.9997} \\ 
\hline
AV / HV & 0.1 & \textbf{2.7841} &  \textbf{0.0062} & \textbf{12.6774} & \textbf{0.9999} \\ \hline
AV / HV & 0.5 & \textbf{2.5417} &  \textbf{0.0123} & \textbf{4.1408} & \textbf{0.9999} \\ 
\hline
\end{tabular}
\end{table}

\par When the null hypothesis of the Markov property test is considered to hold when \(p>0.01\), the average Markov order for the AV trajectory is 1.74s, and for the HV trajectory, it is 3.18s. The average Markov order of HV trajectories is higher than that of AV trajectories. From Table \ref{cats difference}, it can be seen that the \(t\) test results also support this conclusion, with a \(t\) statistic value of 2.7841 and a \(p\) value of 0.0062, indicating that the Markov order of HV trajectories is significantly higher than that of AV trajectories. Additionally, the \(F\) test results further demonstrate that the distribution of the Markov order for HV trajectories is noticeably wider than that of AV trajectories, with an \(F\) statistic value of 12.6774 and a \(p\) value of 0.9999. Besides, it could be found in Table \ref{CATS p_0.01} that, 53.70\% of AV trajectories conform to MP, while in comparison, only 31.15\% of HV trajectories conform to MP. Above evidence indicates that in the CATS Lab ACC dataset, AVs exhibit stronger Markov properties compared to HVs, with a lower Markov order and a higher proportion of MP trajectories. Furthermore, HV trajectories demonstrate greater heterogeneity in their Markov properties compared to AV trajectories.

\par In the cruising scenario, the average Markov order for AV trajectories is 2.47s, while for HV trajectories, it is 4.00s. In the oscillation scenario, the average Markov order for AV trajectories is 1.47s, and for HV trajectories, it is 2.78s. It can be observed that the Markov order for both AV and HV trajectories in the cruising scenario is higher than that in the oscillation scenario. This suggests that when making car-following decisions in the oscillation scenario, AV and HV consider less historical state information compared to the cruising scenario. Additionally, comparing the percentage of MP trajectories, in the cruising scenario, 33.33\% of AV trajectories and 30.00\% of HV trajectories conform to MP, while in the oscillation scenario, 61.53\% of AV trajectories and 31.71\% of HV trajectories conform to MP. This further indicates that the Markov property is stronger for both AV and HV in the oscillation scenario compared to the cruising scenario. Comparing the differences in Markov property between AV and HV in the two scenarios, it can be found that in the cruising scenario, there is no significant difference in the average Markov order between AV and HV trajectories: the \(t\) statistic value is 1.3198, and the \(p\) value is 0.1959. However, the variance of the Markov order distribution for HV trajectories is significantly larger than that of AV trajectories: the \(F\) statistic value is 8.7231, and the \(p\) value is 0.9999. In the oscillation scenario, the Markov order of HV trajectories is significantly higher than that of AV trajectories: the \(t\) statistic value is 2.4476, and the \(p\) value is 0.0166. Moreover, the variance of the Markov order distribution for HV trajectories is significantly larger than that of AV trajectories: the \(F\) statistic value is 26.4571, and the \(p\) value is 0.9999.

\par When the null hypothesis of the Markov property test is considered to hold when \(p>0.05\), the Markov statistical test results are similar to that of \(p>0.01\). The average Markov order for the AV trajectory is 2.46s, and for the HV trajectory, it is 4.13s. The average Markov order of HV trajectories is again higher than that of AV trajectories. The \(t\) statistic results and \(F\) statistical test results shown in Table \ref{cats difference} indicate that the Markov order of HV trajectories is higher than that of AV trajectories (\(t\) statistic value is 2.5417, \(p\) value is 0.0123), and the distribution of the Markov order for HV trajectories is wider than that of AV trajectories (\(F\) statistic value is 4.1408, \(p\) value is 0.9999). Besides, 33.33\% of AV trajectories conform to MP, and 24.59\% of HV trajectories conform to MP. It could be concluded again that AV trajectories exhibit stronger Markov properties compared to HV trajectories, and HV trajectories show greater heterogeneity when comes to Markov properties compared to AV trajectories.

\par In the cruising scenario, the average Markov order for AV trajectories is 3.07s, while for HV trajectories, it is 4.95s. In the oscillation scenario, the average Markov order for AV trajectories is 2.46s, and for HV trajectories, it is 3.71s. It can be concluded that the Markov property is stronger for both AV and HV in the oscillation scenario compared to the cruising scenario. Additionally, there is no significant difference in the average Markov order between AV and HV trajectories: the \(t\) statistic value is 1.6118, and the \(p\) value is 0.1165. However, the variance of the Markov order distribution for HV trajectories is significantly larger than that of AV trajectories: the \(F\) statistic value is 7.8465, and the \(p\) value is 0.9998. In the oscillation scenario, the Markov order of HV trajectories is significantly higher than that of AV trajectories: the \(t\) statistic value is 1.8642, and the \(p\) value is 0.0660. Moreover, the variance of the Markov order distribution for HV trajectories is significantly larger than that of AV trajectories: the \(F\) statistic value is 3.1915, and the \(p\) value is 0.9997.

\par Figure \ref{Fig_catsacc_hist} shows the frequency and probability density of the Markov order for trajectories in the CATS Lab ACC dataset, and Figure \ref{Fig_catsacc_box} shows box plots of the Markov order distribution. Analyzing these two figures, we can draw conclusions similar to the previous ones. In the cruising scenario, there is no significant difference in the Markov order between AV and HV trajectories: in Figure \ref{Fig_catsacc_hist}, the frequency distribution and probability density curves of "AV-C" and "HV-C" are close, and in Figure \ref{Fig_catsacc_box}, the size and position of the boxes in "AV-C, All", "HV-C, All", "AV-C, HOMP", and "HV-C, HOMP" are similar. In the oscillation scenario, the Markov property of AV trajectories is stronger than that of HV trajectories: in Figure \ref{Fig_catsacc_hist}, the frequency distribution of "AV-O" is concentrated on the left side of the horizontal axis, while the frequency distribution of "HV-O" is scattered along the entire horizontal axis. In Figure \ref{Fig_catsacc_box}, the boxes of "AV-O, All" and "AV-O, HOMP" are located lower compared to "HV-O, All" and "HV-O, HOMP", and HV has more outliers, indicating that many HV trajectories exhibit high-order Markov properties. Furthermore, overall, AV trajectories demonstrate stronger Markov properties compared to HV trajectories, and HV trajectories exhibit higher heterogeneity than AV trajectories.

\par To summarize, in the CATS Lab ACC dataset, we selected a total of 54 AV car-following trajectories and 61 HV car-following trajectories. When choosing \(p>0.01\) as the threshold for the null hypothesis to hold in the Markov property statistical test, the average Markov order of AV trajectories is 1.74s, while the average Markov order of HV trajectories is 3.18s. When choosing \(p>0.05\) as the threshold for the null hypothesis to hold in the Markov property statistical test, the average Markov order of AV trajectories is 2.63s, and the average Markov order of HV trajectories is 4.13s. The results of the \(t\) test in both two thresholds indicate that the Markov order of HV trajectories is significantly higher than that of AV trajectories. The results of the \(F\) test show that the variance of the Markov order for HV trajectories is higher than that of AV trajectories. Additionally, in the cruising scenario, there is no significant difference in the Markov property between AV and HV, while in the oscillation scenario, AV exhibits a stronger Markov property compared to HV. Comparing the differences between the two scenarios, both AV and HV have higher Markov properties in the oscillation scenario than in the cruising scenario. This suggests that in the cruising scenario, both AV and HV tend to make car-following decisions based on real-time states, considering less historical information and AV's decision-making process has a stronger dependence on real-time states during such scenarios.

\section{Conclusion}

\par This study introduce a rigorous statistical testing framework to validate Markov properties in vehicle trajectory data. Utilizing datasets from Lyft Level-5 and CATS Lab ACC, we conducted extensive statistical tests to determine if the Markov assumption holds and to what extent it is applicable to different types of driving behaviors.

\par Our findings reveal several key insights:

\par Stronger Markov properties in AVs. AV trajectories exhibit a stronger adherence to the Markov property compared to HV trajectories. A higher percentage of AV trajectories conform to the Markov property, indicating that AVs predominantly base their decision-making processes on the current state rather than on historical states. This reliance on real-time information aligns with the design principles of AV control systems, which emphasize real-time data processing and immediate environmental feedback.

\par Higher Markov orders in HVs. HV trajectories display higher Markov orders on average, suggesting that human drivers incorporate more historical information into their decision-making processes. This is consistent with the cognitive complexity involved in human driving, where decisions are influenced by past experiences, accumulated knowledge, and anticipatory behaviors. The higher Markov orders reflect the nuanced and often non-linear nature of human driving behaviors.

\par Greater variability and heterogeneity in HVs. The variability and heterogeneity in HV trajectories are significantly higher than in AV trajectories. This indicates that human driving behaviors are less predictable and more diverse, influenced by a broader range of factors including individual driving styles, road conditions, and situational awareness. In contrast, AVs demonstrate more uniformity in their decision-making processes, leading to more consistent adherence to the Markov property.

\par Scenario-specific differences. The study also reveals that both AVs and HVs exhibit different Markov properties under varying driving scenarios. For instance, in oscillation scenarios where frequent acceleration and deceleration occur, both AVs and HVs show stronger Markov properties compared to cruising scenarios. This suggests that in dynamic driving conditions, the decision-making processes of both AVs and HVs rely more on the immediate state, possibly due to the increased need for rapid response to changing conditions.

\par This study opens several avenues for future research.
\par Firstly, this research only analyzed the Markov property in car-following behavior trajectories. In real-world driving, there are various complex behaviors, such as lane-changing, turning, overtaking, etc. Analyzing whether the Markov property exists in these trajectories and whether there are differences in the Markov property between AVs and HVs will further contribute to understanding the Markov property of actual driving behaviors.
\par Secondly, analyzing the factors and conditions that influence the Markov property of trajectories is also a potential research direction. This study intends to investigate the differences between Markov and non-Markov trajectories from the perspective of control theory, analyzing them in both time and frequency domains.
\par Another important direction is to explore the implications of these findings for the development of AV control algorithms that can better mimic human driving behaviors, enhancing their acceptance and safety in mixed traffic environments.

\section{Acknowledgments}
\par This work was supported by the National Science Foundation (NSF) under Grant No. 2313578, and No. 2343167. We gratefully acknowledge the support provided by the NSF. Additionally, we acknowledge the support of the Center for High Throughput Computing (CHTC) at University of Wisconsin-Madison \cite{https://doi.org/10.21231/gnt1-hw21}. The distributed high-throughput computing resources provided by the CHTC were invaluable to the completion of all statistical tests presented in this work. We also thank Author Zhou et al. for providing the statistical testing methods and code in their work (\cite{zhou2023testing}), and part of this research was completed using their open-source implementation.

\bibliographystyle{model1-num-names}
% \bibliographystyle{cas-model2-names}

% Loading bibliography database
\bibliography{cas-refs}

% %\vskip3pt

% \bio{}
% Author biography without author photo.
% Author biography. Author biography. Author biography.
% Author biography. Author biography. Author biography.
% Author biography. Author biography. Author biography.
% Author biography. Author biography. Author biography.
% Author biography. Author biography. Author biography.
% Author biography. Author biography. Author biography.
% Author biography. Author biography. Author biography.
% Author biography. Author biography. Author biography.
% Author biography. Author biography. Author biography.
% \endbio

% \bio{figs/pic1}
% Author biography with author photo.
% Author biography. Author biography. Author biography.
% Author biography. Author biography. Author biography.
% Author biography. Author biography. Author biography.
% Author biography. Author biography. Author biography.
% Author biography. Author biography. Author biography.
% Author biography. Author biography. Author biography.
% Author biography. Author biography. Author biography.
% Author biography. Author biography. Author biography.
% Author biography. Author biography. Author biography.
% \endbio

% \bio{figs/pic1}
% Author biography with author photo.
% Author biography. Author biography. Author biography.
% Author biography. Author biography. Author biography.
% Author biography. Author biography. Author biography.
% Author biography. Author biography. Author biography.
% \endbio

\end{document}